%% file: root.tex
\documentclass[journal]{IEEEtran}
\usepackage{amsmath,amsfonts}
\usepackage{algorithm} 
\usepackage[pagebackref,breaklinks,colorlinks,citecolor=green]{hyperref}
\usepackage{array}
\usepackage[caption=false,font=normalsize,labelfont=sf,textfont=sf]{subfig}
\usepackage{url}
\usepackage{textcomp}
\usepackage{stfloats}
\usepackage{url}
\usepackage{verbatim}
\usepackage{graphicx}
\usepackage{cite}
\usepackage{amssymb}
\usepackage{latexsym}
\usepackage{float}
\usepackage{indentfirst}
\usepackage{multirow} 
\usepackage{algpseudocode}  
\usepackage{booktabs}

\usepackage{soul,color}
\usepackage{xcolor}
{}
{}
{}

\usepackage{algorithmicx}
\usepackage{algpseudocode}

\begin{document}

\title{PNAS-MOT: Multi-Modal Object Tracking with Pareto Neural Architecture Search}

\author{Chensheng Peng\textsuperscript{\rm 1}, Zhaoyu Zeng\textsuperscript{\rm 2}, Jinling Gao\textsuperscript{\rm 2}, Jundong Zhou\textsuperscript{\rm 2} \\
Masayoshi Tomizuka\textsuperscript{\rm 1}, 
Xinbing Wang\textsuperscript{\rm 2}, Chenghu Zhou\textsuperscript{\rm 2}, Nanyang Ye\textsuperscript{\rm 2} 
\smallskip
\\ 
\textsuperscript{\rm 1} UC Berkeley \quad \textsuperscript{\rm 2} Shanghai Jiao Tong University
\thanks{Nanyang Ye is the corresponding author.}
}



\maketitle

\begin{abstract}
Multiple object tracking is a critical task in autonomous driving. Existing works primarily focus on the heuristic design of neural networks to obtain high accuracy. As tracking accuracy improves, however, neural networks become increasingly complex, posing challenges for their practical application in real driving scenarios due to the high level of latency. In this paper, we explore the use of the neural architecture search (NAS) methods to search for efficient architectures for tracking, aiming for low real-time latency while maintaining relatively high accuracy. Another challenge for object tracking is the unreliability of a single sensor, therefore, we propose a multi-modal framework to improve the robustness.  Experiments demonstrate that our algorithm can run on edge devices within lower latency constraints, thus greatly reducing the computational requirements for multi-modal object tracking while keeping lower latency. Code is available at \url{https://github.com/PholyPeng/PNAS-MOT}.
\end{abstract}


\section{Introduction}\label{sec1}
\IEEEPARstart{M}{ultiple} object tracking (MOT) is a fundamental task of consistently assigning a unique ID to each observed object within a video sequence, which holds significant importance across various domains, including motion planning, safe robot navigation, and autonomous driving \cite{liu2023regformer}. The primary challenge inherent to MOT lies in establishing precise associations between tracklets from preceding frames and the object detections within the current frame. To tackle the complexities of multi-object tracking, two main-stream paradigms have emerged: tracking-by-detection \cite{zhang2019robust, wang2023interactive} and joint-tracking-and-detection \cite{voigtlaender2019mots, feichtenhofer2017detect, vo2012multi}. The tracking-by-detection paradigm follows a two-stage process. Initially, a pre-trained detector is employed to procure object detections, after which a tracker undertakes the data association task, assigning a distinct ID to each detected object across successive frames. On the other hand, the joint-tracking-and-detection paradigm endeavors to achieve detection and tracking concurrently, leveraging the benefits of joint optimization strategies. In this paper, our focus lies specifically on the tracking aspect, so we adopt the tracking-by-detection approach due to its its inherent efficiency and proven effectiveness in addressing the complexities of object tracking tasks.

In the task of Multiple Object Tracking, two primary sensor modalities, namely images and LiDAR point clouds, are extensively utilized. Various methodologies have been devised to address MOT challenges leveraging these sensor modalities \cite{yin2021center, wang2023interactive, bergmann2019tracking, weng20203d, frossard2018end}. Recent studies have revealed the limitations of relying solely on a single modality, often resulting in mismatching issues. For example,
cameras offer the advantage of capturing rich texture details but are susceptible to variations in lighting conditions. On the other hand, LiDAR sensors excel in capturing 3D geometry information and maintaining robustness in adverse weather conditions, yet with limitations on perceiving distant object detection owing to the sparsity of points.
As a result, methods \cite{sharma2018beyond,  weng2020gnn3dmot, kim2021eagermot} utilizing the fusion features have been proposed. However, the improvement of accuracy from multi-modal fusion often comes at the cost of significant latency and high energy consumption. These factors present significant challenges when deploying such methods in practical scenarios, particularly within autonomous driving systems. 

To address these challenges, we propose the integration of neural architecture search (NAS) techniques to search for efficient deep neural networks (DNNs) \cite{elsken2019neural}.
Early NAS methods primarily focused on improving accuracy by searching for network architectures within an expansive search space containing numerous structures. For example, DARTs \cite{liu2018darts} propose to formulate the searching task in a differentiable manner,  focusing solely on the search for minimal network blocks, which are subsequently utilized to construct complete networks. Under the context of the challenging multi-modal MOT, we focus on latency-constrained NAS to search for architectures that exhibit superior performance while maintaining minimal latency. Specifically, we propose a Pareto optimization scheme to find the optimal Pareto frontier for the best trade-off between latency and accuracy. Subsequently, we select the network architectures that are capable of simultaneously achieving the desired levels of latency and accuracy. Moreover, our proposed multi-modal framework integrates information from both sensors to enhance robustness in handling challenging scenarios. Through a learned weighted feature fusion mechanism, the framework assigns greater significance to LiDAR point cloud features when cameras encounter difficulties or fail, and vice versa.

To summarize, our contributions are as follows:\begin{itemize}
\item We propose a constrained neural architecture search (NAS) method that searches for network architectures capable of completing the MOT task within a specified time limit. This is solved by a Pareto frontier searching algorithmic scheme, which finds a suitable latency accuracy trade-off.
\item We evaluate the proposed algorithmic framework on the KITTI benchmark and achieve 89.59\% accuracy close to the SOTA methods while keeping the latency below 80 milliseconds on different edge devices.
\end{itemize}
 

\section{Related Work}
\subsection{Multiple Object Tracking}

Recent MOT methods have made remarkable progress largely due to powerful deep neural networks. Sun \textit{et al.} \cite{sun2019deep} concentrate on object affinity between different frames and propose an efficient online tracking network named Deep Affinity Network (DAN). Wang \textit{et al.} \cite{wang2023interactive} introduce an attention mechanism to fuse features from multiple modalities. However, object occlusion and overlapping significantly impair the accuracy of localizing and tracking objects. To overcome this issue, Ren \textit{et al.} \cite{ren2017accurate} introduce the Recurrent Rolling Convolution (RRC) architecture. Tracklet-Plane Matching (TPM), proposed by Peng \textit{et al.} \cite{peng2020tpm}, is an approach to reduce the influence of noisy or confusing object detection and improve the performance of MOT. 

Existing work \cite{cui2022exploiting}, \cite{wang2023implicit} propose methods for more efficient use of point cloud information for better object tracking performance, and \cite{li2023autonomous} propose methods for improving the robustness of tracking. Wu \textit{et al.} \cite{wu20213d} utilize prediction confidence to guide data association and build a more robust tracker for objects temporarily missed by detectors. Dunnhofer \textit{et al.} \cite{dunnhofer2021weakly} propose a weakly-supervised adaptation strategy and utilize knowledge distillation to overcome inadequate tracking accuracy in many domains due to distribution shift and overfitting. Jong \textit{et al.} \cite{de2022apple} introduce APPLE MOTS, a dataset of homogeneous objects and propose TrackR-CNN and PointTrack architecture for joint detection and tracking. Fu \textit{et al.} \cite{fu2023scale} propose a novel scale-aware domain adaptation framework, ScaleAwareDA, to solve the gap issue on object scale between the training and inference phases. 

\subsection{Multi-modal fusion}
Multi-modal fusion remains a burgeoning area of exploration within the domain of autonomous driving \cite{peng2023delflow, xie2023sparsefusion}. By combining features derived from diverse modalities, it offers a potent strategy to mitigate the limitations inherent in single-modality approaches in Multiple Object Tracking (MOT) tasks, including issues related to mismatching and unreliability. Addressing the challenge of weak-pairing characteristics in multi-modal fusion, Liu \textit{et al.} \cite{liu2017weakly} have made notable strides in enhancing the fusion effectiveness. A variety of modalities and sensors are employed for multimodal fusion regarding the application scenarios. For instance, Qu \textit{et al.} \cite{qu2017active} utilize visible, infrared, and hyperspectral sensors simultaneously to increase accuracy and robustness for object detection and tracking tasks.
In a similar way, Zhang \textit{et al.} \cite{zhang2018vehicle} leverage data extracted from paired images and velocities, and then propose an efficient vehicle tracker, underscoring the versatility of multi-modal fusion techniques. EagerMOT, introduced by Kim \textit{et al.} \cite{kim2021eagermot}, exemplifies the integration of information from depth sensors and cameras, facilitating the identification and localization of distant incoming objects with heightened precision and reliability.
Moreover, Xu \textit{et al.} \cite{xu2021fusionpainting} delve into the semantic-level fusion of 2D RGB images and 3D point clouds, aiming to enhance detection performance through a nuanced integration strategy. 

\subsection{Neural Architecture Search}
Neural Architecture Search (NAS) is a method for automated neural network search particularly in pursuit of high accuracy. NAS is usually implemented with two categories of approaches,\textit{ i.e.}, reward-based methods (\textit{e.g.}, Reinforcement Learning) and gradient-based methods. Zoph \textit{et al.} \cite{zoph2016neural} utilize reinforcement learning to train an RNN network that generates the model descriptions of neural networks. Liu \textit{et al.} \cite{liu2018darts} introduce a differential search space that allows the use of gradient descent for architecture search. Constrained-NAS are proposed to satisfy relatively insufficient computing resources for limited computing platforms (\textit{e.g.}, Edge-GPU, FPGA). 
Tan \textit{et al.} \cite{tan2019mnasnet} introduce latency constraints into the mobile neural architecture search (MNAS) approach and achieve high accuracy on mobile equipment. However, a trade-off between resource consumption and accuracy deteriorates the network accuracy and fails to meet the constraints strictly. To address this, Nayman \textit{et al.} \cite{nayman2021hardcore} further strengthens the latency constraints and proposes the HardCoRe-NAS method, which satisfies the constraint tightly without sacrificing accuracy.

\section{Methods}

\subsection{Problem Statement}

In this paper, we follow the widely adopted tracking-by-detection paradigm, where a set of object detections is first obtained from a pre-trained detector. Let the detection results in $T$ consecutive frames of a video sequence be $\mathcal{O}=\{O^1, \cdots O^{t}, \cdots O^{T}\}$, where $O^t = \{o_j^t\}, j = 0, 1, \cdots N^t$, and $o_j^t = (b_j^t,t)$. $N_t$ is the total number of detections of the $t$-th frame, and $b_j^t$ is usually a 2D bounding box on image plane for 2D detection results, while a 3D bounding box for 3D detection results, and $t$ is the time stamp. The goal of multiple object tracking is to assign an ID for each instance detection $o_j^t$. A tracklet is defined as a set of object detections in different frames, $\tau = \{o_{k_1}^{t_1},o_{k_2}^{t_2},o_{k_3}^{t_3}, \cdots \}$. The objective of consistent ID assignment can be interpreted as finding a set of tracklets $\mathcal{T} = \{\tau_i\}, i = 0,1, \cdots n$, that can best explain the object detections. The problem is defined as predicting the correct tracklets $\mathcal{T}$ by maximizing the conditional probabilities given the set of object detections $\mathcal{O}$.
\begin{equation}
    \mathcal{T^{\ast}} = \mathop{\arg\max}_{\mathcal{T}}\mathbb{P}\left(\mathcal{T}\mid\mathcal{O}\right).
\end{equation}

\begin{figure}[t]
    \centering
    \includegraphics[width=\linewidth]{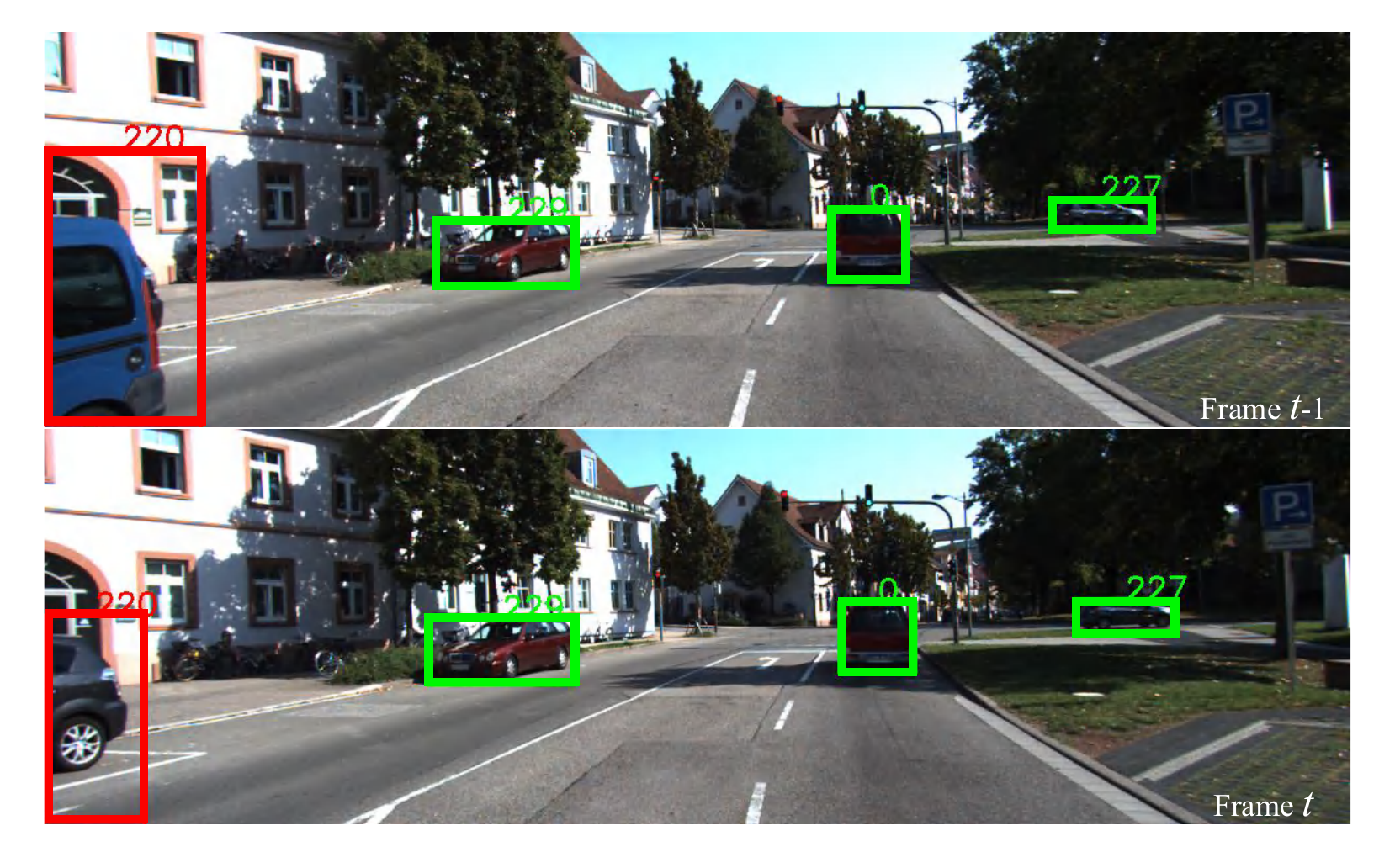}
    \caption{Successful and failed examples of multi-object tracking. \textbf{Green bounding box:} The assigned IDs 0, 227, and 229 are consistent in two consecutive frames $t-1$ and $t$, which are successful trackings. \textbf{Red bounding box:} The ID 220 is wrongly assigned to the black vehicle in frame $t$, which is a failure tracking. Under the tracking-by-detection setting, Multi-object tracking can be regarded as a downstream detection task. It is hard to track dynamic objects under occlusion, which is common in real driving scenarios and significant for safe autonomous driving.}
    \label{fig:demo}
\end{figure}

For a successful tracking process, the ID assignment must be consistent and unique for the same instance in a video sequence. As shown in Figure \ref{fig:demo}, the three tracked objects whose IDs are 0, 227, and 229 respectively, are a successful tracking procedure, because the assigned IDs are consistent in two consecutive frames $t-1$ and $t$. However, for the car whose ID is 220 in frame $t-1$, the ID is assigned to the blue vehicle at the left bottom of the image, but it's wrongly assigned to the black vehicle in frame $t$.


\begin{figure*}[t]
    \centering    \includegraphics[width=\textwidth]{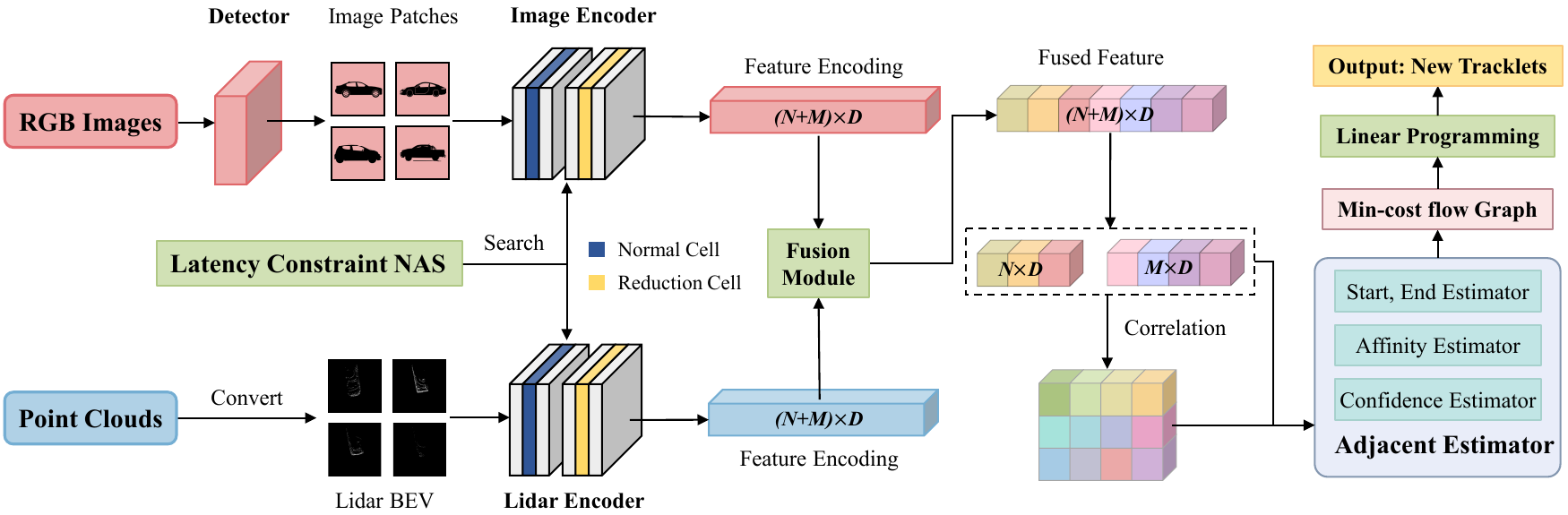}
    \caption{The structure overview of the tracking framework. We take image patches and LiDAR point clouds as input, which are processed respectively by a corresponding encoder. A fusion module is then used to fuse the multi-modal features. The fused features are split into \textbf{tracklet features} and \textbf{detection features}, from which the \textbf{correlation features} can be calculated. These features are input to the following adjacent estimator to infer confidence and affinity scores. At last, linear programming is used to update the new tracklets from the inferred scores.}
    \label{fig:struct}
\end{figure*}

\subsection{Tracking Network Structure}
\label{sec:trackingnet}
For the tracking network, we take the image and point clouds as input. Through a detector, we obtain the object detections in frame $t$. To reduce the amount of data, we crop the instance detections from the full image and then resize them to image patches whose size is 256 $\times$ 256. In 3D spaces, we remove the point clouds outside the bounding box and convert the remaining points to Birds Eye View (BEV) images whose resolution is 256 $\times$ 256.

Next, as shown in Figure \ref{fig:struct}, in an online setting, tracklets $T_{t-1} = \{ o_{n}^{t-1}\}$, where $n = 1, 2, \cdots, N$, with $N$ being the number of tracklets in frame $t-1$ and object detections $O^t = \{ o_{m}^{t}\}$, where $m = 1, 2, \cdots, M$, with $M$ being the number of detections in frame $t$, are fed to the feature extraction network, which consists of two branches, an image encoder, and a LiDAR encoder, each comprised of a normal cell and a reduction cell searched from the latency constraint NAS method. We can get the feature encoding from each modality and then add them together to obtain the fused feature, whose size is $(N+M) \times D$.  

During the following correlation process, the fused feature is split to a $N \times D$ vector and a $M \times D$ vector. Then the two vectors are used to calculate the correlation features between existing tracklets and current detections. In this work, the difference calculation is defined as 
$ \textbf{D}_{n,m} = \mid f_n^{t-1}-f_m^{t}\mid,$
where $f_n^{t-1}$ and $f_m^{t}$ are the features of the $n$-th object in frame $t-1$ and the $m$-th in frame $t$ respectively. 

Following previous works \cite{frossard2018end,lenz2015followme,schulter2017deep,zhang2019robust}, MOT can be solved as a min-cost flow problem. Together with the fused features, correlation features are then fed to the adjacency estimator adapted from \cite{zhang2019robust}, which infers the corresponding new, end, confidence, and affinity scores. The predicted scores are used to construct a graph with each edge representing the cost.
 Then, we use the mixed integer programming (MIP) solver provided by Google OR-Tools \footnote{https://developers.google.com/optimization} to find the optimal solution in the min-cost graph, generating the updated tracklets by adding the connected detections to their corresponding tracklets or initiating a new tracklet.

During the data association process, for the matched tracklets and detections, the tracklet will be updated by appending the detection to the tracklet. The object detection will obtain the ID from the corresponding tracklet through the ID propagation process. For the unmatched detection, it will not be instantly initialized as a new tracklet, instead, a birth check procedure is used to check whether it's a wrong detection or not. Only when the detection appeared in $t_{birth}$ consecutive frames, the detection will be initialized. Similarly, for the unmatched tracklets, we won't delete them immediately, instead, we check if they disappeared in $t_{death}$ frames.

\subsection{Hardware latency function}
To enable efficient NAS, we need to estimate the latency of the network architecture on specific hardware platforms. We test the latency of different operations. We first list all the possible $in\_channels, out\_channels, resolutions$, and other parameters used in our architectures. Every set of parameters is defined as a configuration. We set a random input with the corresponding input size of every configuration and pass it to the operation. The time spent is recorded by a preset timer. After hundreds of repetitions of such processes and an averaging procedure, we obtain a dictionary of the average latency per iteration for every operation with different configurations.

The architecture encoding $\alpha_i$ for each operation $\mathop{op_i}$, where $i = 1,2, \cdots n$ with $n$ being the number of operations, can be obtained through a Softmax function, then the estimated latency can be obtained from a weighted sum:
$$
\operatorname{Lat}(\alpha) = \sum_{i=1}^{n} \alpha_i \cdot \operatorname{Lat}(\mathop{op_i}),
$$
where the $\operatorname{Lat}(\mathop{op_i})$ can be easily acquired through a table-lookup process. 
\subsection{Pareto Optimization}
\begin{figure}[t]
    \centering
    \includegraphics[width=\linewidth]{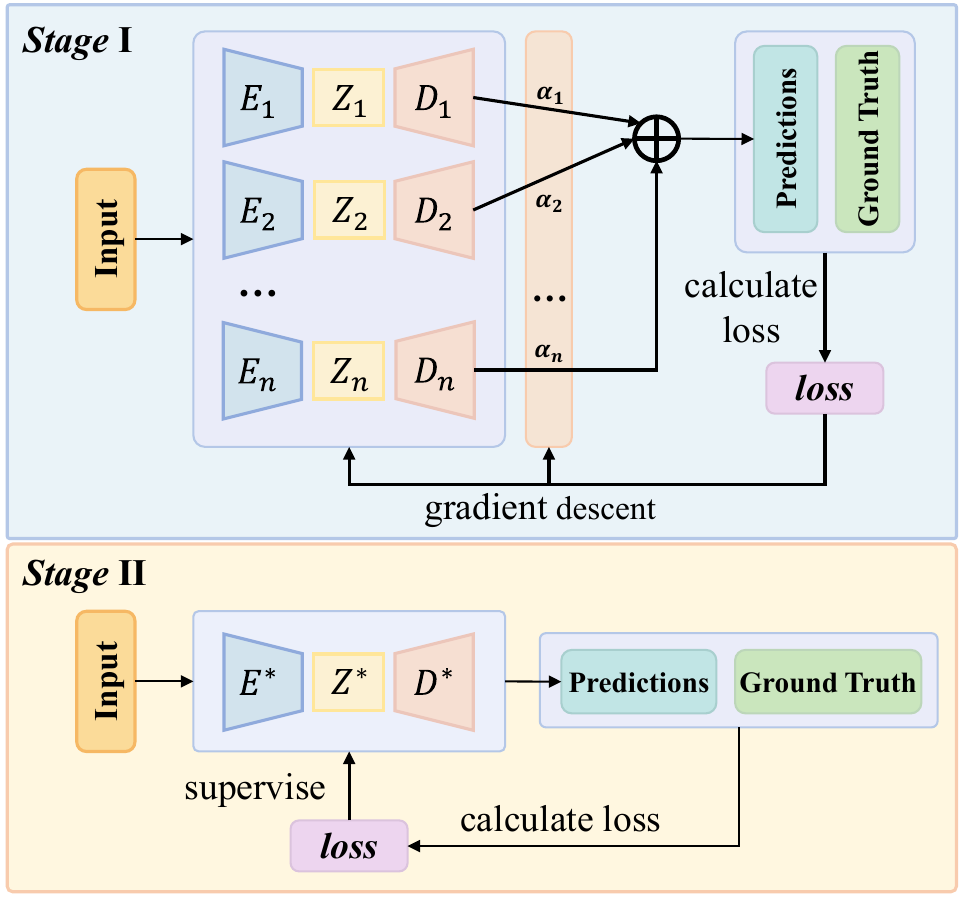}
    \caption{The searching and training process is divided into two stages. We obtain the optimal network structures in the first stage through constrained NAS and get the optimal network parameters in the second training stage.}
    \label{fig:alg}
\end{figure}

Fig. \ref{fig:alg} shows the structure overview of the MOT framework. We aim to find an optimal DNN that suits the latency accuracy trade-off. To achieve this trade-off, we design a two-stage Pareto optimization scheme. 

For the first stage, we search for efficient network structures with the proposed constrained-NAS method. The backbone architecture to be searched is parameterized as $\alpha$, and the search space is denoted as $\mathcal{S}$. Therefore, the complete network parameters are $\zeta = (\alpha, \theta)$, consisting of the architecture $\alpha$ and the corresponding weights $\theta$. Thus, the optimization problem is formulated as follows with tracking accuracy and latency objectives,
\begin{equation}
\label{eq:stage11}
\begin{split}
&\min_{\alpha \in \mathcal{S}, \theta} \{\mathbb{E}_{\mathcal{O}, \mathcal{T} \sim \mathcal{D}_{val}}\left[\mathcal{L}\left(\mathcal{O}, \mathcal{T} \mid {\theta}, \alpha\right)\right]\} 
\end{split}
\end{equation}

where $\mathcal{D}_{val}$ denotes the distribution of the validation dataset, and $\mathcal{L} = \mathcal{L}_{track} + \lambda\mathcal{L}_{latency}$ denotes the overall loss function which consists of the performance loss $\mathcal{L}_{track}$, the latency function $\mathcal{L}_{latency}$ and the weight coefficient $\lambda$. To solve this multi-objective optimization problem, we implement a Pareto optimization procedure, consisting of two alternating stages.

Our goal for the first stage is to find the best architecture $\alpha^{*}$ of the backbone on the training dataset $\mathcal{D}_{train}$.
\begin{equation}
\label{eq:stage12}
\alpha^{*}=\underset{\alpha}{\operatorname{arg min}}\ \mathbb{E}_{\mathcal{O}, \mathcal{T} \sim \mathcal{D}_{train}} \left[\mathcal{L}\left(\mathcal{O}, \mathcal{T} \mid \theta, \alpha \right)\right] \\
\end{equation}
Note that we set $\lambda > 1$ since we mainly focus on reducing the latency in the first stage. We avoid solely using the latency loss function because $\alpha$ can influence both latency and performance.


For the second stage, we attempt to find the best model $\zeta^{*}$ with the highest accuracy using the searched architecture $\alpha^{*}$ from stage \uppercase\expandafter{\romannumeral1}. The optimization problem can be formulated as,
\begin{equation}
\label{eq:stage2}
\min_{\theta} \ \mathbb{E}_{\mathcal{O}, \mathcal{T} \sim \mathcal{D}_{val}}\left[\mathcal{L}_{track}\left(\mathcal{O}, \mathcal{T} \mid \theta,  \alpha^{*}\right)\right] \\
\end{equation}

After several iterations of the two-stage searching on the Pareto front of latency and accuracy, we can obtain the optimal parameters $\zeta^{*}$.
\begin{equation}
\begin{split}
&\theta^{*}=\underset{\theta}{\operatorname{arg min}}\ \mathbb{E}_{\mathcal{O}, \mathcal{T} \sim \mathcal{D}_{train}} \left[\mathcal{L}\left(\mathcal{O}, \mathcal{T} \mid \theta, \alpha^{*} \right)\right] \\
& \zeta^{*} = (\alpha^{*}, \theta^{*})
\end{split}
\end{equation}

\subsection{Two-stage Neural Architecture Search}

\begin{algorithm}[t]  
  \caption{Stage 1: search for feature extraction backbone 
  }  
  \begin{algorithmic}
    \Require  
    Object detections $O$, Ground truth tracklets $\mathcal{T}$
    \Ensure  
    Optimal network structure encoding $\alpha^{*}$
    \State Initialize architecture $\alpha = \alpha_0$ 
      \For{each epoch $i$}
      \If {converged}  
      \State Generate the optimal architecture encoding $\alpha^{*}$;
      \Else
      \State Update the architecture encoding $\alpha$ 
      \State Calculate the estimated latency from $\alpha$
      \For {each iteration $t$}
      training 
      \State backpropagation with loss function \State $\mathcal{L}_{track}+\lambda\mathcal{L}_{latency}$
      \State Update the network parameters $\theta$
      \EndFor
      \State Evaluation on the validation dataset
      \EndIf
   \EndFor  
  \end{algorithmic}  
\label{alg:stage1}  
\end{algorithm}

We divide the training process into two stages. In the first stage named \textbf{train search}, we first update the architecture parameters $\alpha=\alpha_0$. Then for each epoch, we use the architecture from the last epoch to build our tracking network. We train the tracking network for $T$ iterations, after which the model will be evaluated on a validation dataset to test its performance. If the result does not converge, we continue to update the architecture parameters $\alpha$ for the next training epoch. In order to search for an efficient network structure with a low inference time, we add a latency constraint to the loss function for the first stage.

The second stage is the further training stage. In stage one, the operations between each node are mixed operations, each with a different weight ranging in $(0,1)$, which causes the model will be extremely heavy. Therefore, we will remove the operations with low weight to reduce the size of the model. Pruning the low-weight operation will lead to a gap between the model from the first stage and the second stage. As a result, we continue to train the pruned model to get the best tracking network. In the second stage, the latency constraint is removed because the architecture will not be updated anymore. We focus on the improvement of accuracy in this stage.

\begin{algorithm}[htb]
  \caption{Stage 2: Training with the searched architecture}
  \label{alg:stage2}  
  \begin{algorithmic}
    \Require  
    Object detections $O$, Ground truth tracklets $\mathcal{T}$
    \Ensure  
    Optimal network weights $\theta^{*}$
    \State Initialize architecture $\alpha = \alpha^{*}$ 
    \For{each iteration $t$} 
      \If {not converged}  
      \State Backpropagation using loss function $\mathcal{L}_{track}$
      \State Update the weights $\theta$ for the tracking network 
      \Else
      \State Save the model $\zeta^{*}$ with best performance
      \EndIf
      \State Evaluation on validation dataset at a fixed interval
      \EndFor
  \end{algorithmic}  
\end{algorithm}



\section{Experiments}

\begin{figure}[t]
    \centering
    \includegraphics[width=\linewidth]{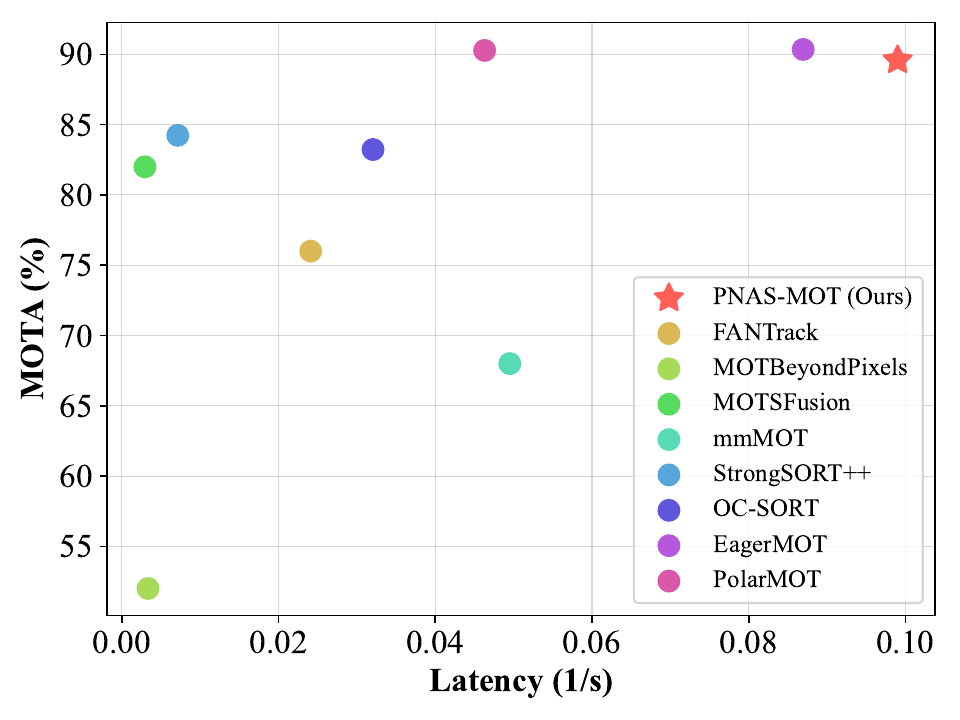}
    \caption{Comparison with other methods on the KITTI testing dataset. The x-axis represents the reciprocal of the latency. As shown in the figure, our method minimally sacrifices the MOTA metrics and achieves outstanding performance on latency, with less than half the latency of mmMOT.}
    \label{fig:res}
\end{figure}

\begin{figure*}[!ht]
    \centering
    \includegraphics[width=\textwidth]{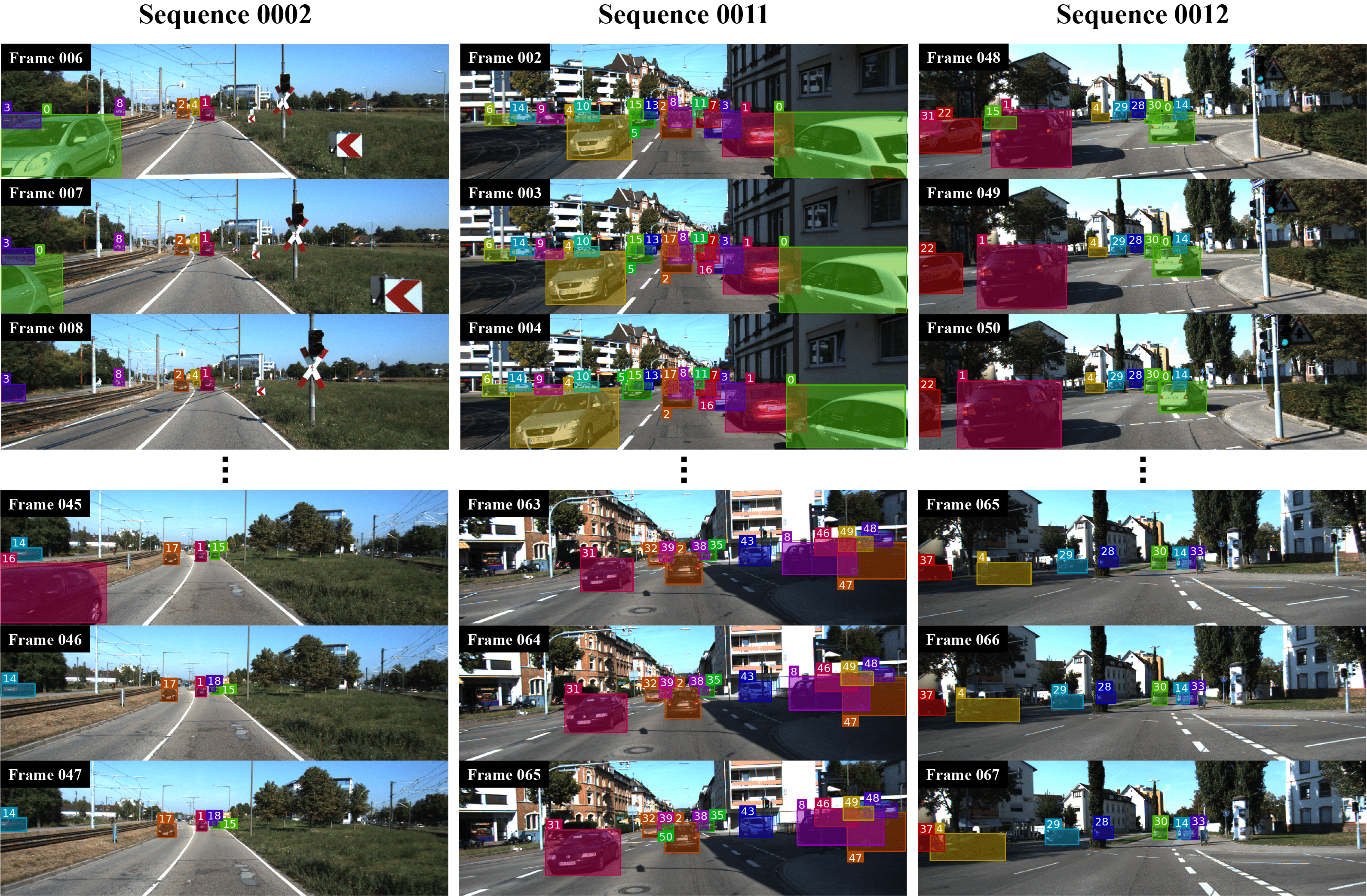}
    \caption{Tracking results of our multi-object tracking model. We illustrate the tracking results of different sequences from the testing dataset of the KITTI benchmark. These are some of the extreme conditions under highly complex environmental challenges. Yet our model accurately keeps track of every object despite the interference of illumination, shadows, and occlusions. This suggests our multi-object tracking model has high accuracy and robustness concerning complex background environments.}
    \label{fig:qual}
\end{figure*}

    
\subsection{Accuracy on KITTI Benchmark}
Our model is trained and evaluated on the KITTI Benchmark \cite{geiger2012we}, which consists of 29 testing sequences and 21 training sequences. Since KITTI does not provide an official validation dataset, we choose sequences 00, 02, 05, 07, 10, 11, 14, 16, 17, 18, and 19 as the validation dataset and the rest as the training dataset. 

We submit the tracking results obtained from our model with the best performance on the validation dataset to the KITTI server, to evaluate our performance on the testing dataset. The comparison of runtime and accuracy can be observed in Figure~\ref{fig:res}. We compared our performance with other methods \cite{baser2019fantrack, sharma2018beyond, luiten2020track, zhang2019robust, du2023strongsort, cao2023observation, kim2021eagermot, kim2022polarmot}. Though our method may not come state of the art in accuracy (MOTA), our model can maintain relatively high accuracy with a short inference time. The quantitative results are demonstrated in Table~\ref{table:res}, and the qualitative results are illustrated in Figure~\ref{fig:qual}.

\input{tables/kitti_res}


\subsection{Latency measured on different devices}

\input{tables/latency_table}

We randomly chose sequence 19 from the KITTI dataset as a validation set for our evaluation, containing 1059 data, images, and point cloud files. We use the validation data to calculate the average inference time.

As Table~\ref{tab:latency} shows, we evaluated our algorithm on different GPUs, including Jetson Nano, Jetson Orin Nano, GTX, Quadro, and TITAN. We conduct experiments on both edge devices and effective GPUs. The results show that our algorithm can run on Jetson Nano, whose memory size is only 2GB, and the inference time is under 80ms. When running on other high-performance GPUs, the inference time can be reduced to less than 8ms at most.

During the search process, we made a trade-off between accuracy and latency. As can be observed from Table~\ref{tab:mota_latency}, we can relax the latency constraint by reducing the parameter $\lambda$ in the loss function of the first stage, resulting in a higher accuracy with higher latency.

\input{tables/MOTA_latency}

\subsection{Ablation studies}

\input{tables/ablation_search_mode}
To evaluate the effectiveness of the latency constraint neural architecture search method. We conduct an ablation study on the KITTI dataset \cite{geiger2012we}. We compare the performance of three different search modes. For search mode 0, we only search for the image branch of the feature extraction backbone, and we use ResNet18 as the LiDAR branch. Similar to search mode 1, we use ResNet34 as the image branch and solely search for the LiDAR branch. Besides, we search for both image and LiDAR branches simultaneously under the setting of search mode 2. The experiments are conducted on Quadro RTX 6000. We choose sequence 0,2,5,7,10,11,14,16,18,19 of the KITTI dataset to evaluate the MOTA and sequence 19 to evaluate the latency of our searched network architectures.

As shown in Table~\ref{tab:search}, compared with search mode 0 and 1, at search mode 2, the latency can be reduced significantly, and the MOTA can be improved, especially the memory size during inference can be reduced by 22.3\% and 14.8\%, respectively. The model trained from search mode 2 can run on Jetson Nano, while the other two cannot since the Jetson Nano only has a 2GB memory size.

\subsection{Implementation Details}
We test the latency of our model on different devices. The best model is trained on an RTX 8000, and the batch size we use is 1. We use ADAM as our optimizer with a learning rate of $3e^{-6}$. 
The search space is adapted from DARTs \cite{liu2018darts} while modified to fit the MOT tasks. The candidate operations are as follows:{ none, identity, 3 $\times$ 3, 5 $\times$ 5, 7 $\times$ 7 separable convolutions, 3 $\times$ 3, 5 $\times$ 5 dilated convolution, 3 $\times$ 3 max pooling, and 3 $\times$ 3 average pooling}.

In our designed architecture, there are two branches in the feature extraction backbone, one is image modality, and the other one is LiDAR modality. For each branch, the network consists of two different kinds of cell, normal cell and reduction cell, where the former one does not change the feature channels while the latter one does. Each cell consists of $N$ nodes, where each edge between two nodes represents an operation in a pre-defined search space. Our goal is to find the best architecture of normal and reduction cells, including the connection between each node and the operation of each edge. Notably, the searched architecture parameters $\alpha$ are shared by cells of both branches.

\section{{Limitation}}
We employ DARTs \cite{liu2018darts} as the backbone for Neural Architecture Search, where the network structure is parameterized by weights assigned to edges between nodes. The searched structure is obtained by discarding edges with weights falling below a certain threshold. Consequently, the structure of Stage II deviates slightly from that of Stage I. This discrepancy introduces a minor deviation in the searched structure from the optimal configuration.

\section{Conclusion}
In this paper, we introduce a latency-constrained multiple modalities fusion neural architecture search method for MOT tasks. Numerical experiments have demonstrated the superiority of our proposed scheme. We have achieved 89.59\% accuracy close to the SOTA methods while keeping the latency below 80 milliseconds. This methodology may serve as a foothold for future efficient autonomous driving research.

\bibliographystyle{plain} 
\bibliography{root}

\vfill

\end{document}

%% file: tables/kitti_res.tex
\begin{table*}[htpb]
  \caption{Comparison on the testing datasets of KITTI.}
  \label{table:res}
  \centering
  \begin{tabular}{l|ccccccccccc}
    \toprule
    Method & MOTA$\uparrow$ & MOTP$\uparrow$  & IDSW$\downarrow$ & HOTA$\uparrow$ & DetA $\uparrow$ & AssA $\uparrow$ &  FN$\downarrow$ & FP$\downarrow$  & Frag$\downarrow$ & MT$\uparrow$ & ML$\downarrow$\\
    \midrule
    DSM \cite{frossard2018end} &  73.94 & 83.5  & 939  & 60.05 & 64.09 & 57.18 & 637  & 7388 & 737      & 59.38 & 8.46  \\
    extraCK \cite{gunduz2018lightweight} & 79.29 & 82.06 & 520  & 59.76 & 65.18 & 55.47 & 675  & 5929 & 750 & 62.31 & 5.85  \\
    FANTrack \cite{baser2019fantrack} & 75.84 & 82.46 & 743 & 60.85 & 64.36 & 58.69 & 1305 & 6262 & 701 & 62.77 & 8.77 \\
    MOTBeyondPixels \cite{sharma2018beyond} &  82.68 & {85.50}  & 934  & 63.75 & 72.87 & 56.4  & 741  & 4283 & 581 & 72.61 & 2.92  \\
    PMBM \cite{scheidegger2018mono} & 79.23 & 81.58 & 485  & 59.12 & 65.43 & 54.28 & 1024 & 5634 & 554  & 62.77 & 6.46  \\
    JCSTD \cite{tian2019online} & 80.24 & 81.85 & \textbf{173 } & 65.94 & 65.37 & \textbf{67.03} & \textbf{405}  & 6217 & 700 & 57.08 & 7.85  \\
    mmMOT \cite{zhang2019robust} & 83.23 & 85.03 & 733  & 62.05 & 72.29 & 54.02 & 752  & 4284 & 570  & 72.92 & 2.92  \\
    \midrule
    Ours & \textbf{89.59} & \textbf{85.44} & 751 & \textbf{67.32} & \textbf{77.69} & 58.99 & 568 &\textbf{ 2261} & \textbf{276} & \textbf{86.59} & \textbf{2.46} \\
    \bottomrule
  \end{tabular}
\end{table*}

%% file: tables/latency_table.tex
\begin{table*}[!ht]
\caption{Latency on different devices.  The performances of our algorithm on different devices strictly satisfy latency requirements and are far better than the minimum limit.\label{tab:latency}}
\centering
{\begin{tabular}{lcccc}
\toprule
Devices & Latency (milliseconds) & Floating-point performance (GFLOPS) & compute capability & Memory Size (GB)\\
\midrule
Jetson Nano & 78 & 235.8 & 5.3 & 2\\
Jetson Orin Nano & 58 & 1280 & 8.7 & 8\\
GTX 1050 Ti & 18 & 2138 & 6.1 & 4\\
GTX 2080 Ti & 10 & 13450 & 7.5 & 11\\
Quadro RTX 6000 & 8 & 16310 & 7.5 & 24\\
Quadro RTX 8000 & 8 & 16310 & 7.5 & 48\\
TITAN RTX &  9 & 16310 & 7.5 & 24\\
TITAN V & 8 & 14900 &7.0 & 12\\
\bottomrule
\end{tabular}}{}
\end{table*}

%% file: tables/MOTA_latency.tex
\begin{table}
  \centering
  \caption{MOTA-latency trade-off on the validation dataset}
  \label{tab:mota_latency}
  \begin{tabular}{ccccc}
    \toprule
   $\lambda$ & Latency (ms) & MOTA (\%) &  HOTA (\%) &  IDSW \\
    \midrule
   10.0 & 6.2 & 89.42 & 71.02 & 334\\
  1.0 &  8.3 & 90.48 & 73.85 &  235\\
  0.1 &  10.1 & 90.91 & 75.84 & 188 \\
   0.01 & 21.0 & 91.01 & 76.82 & 183\\
  \bottomrule
\end{tabular}
\end{table}

%% file: tables/ablation_search_mode.tex
\begin{table}[!htbp]
  \caption{Comparison between different search mode}
  \label{tab:search}
  \begin{tabular}{cccc}
    \toprule
    Search Mode & Latency (ms) & MOTA \% & Memory Size (MB) \\
    \midrule
    0  & 12 & 90.3 & 2462 \\
    1 & 10 & 90.6 & 2246 \\
    2  & \textbf{8} & \textbf{90.9} & \textbf{1913}\\
  \bottomrule
\end{tabular}
\end{table}